\documentclass[conference]{IEEEtran}
\IEEEoverridecommandlockouts
\usepackage{cite}
\usepackage{amsmath,amssymb,amsfonts,amsthm}
\usepackage{graphicx}
\usepackage{textcomp}
\usepackage{xcolor}
\usepackage{tcolorbox}
\usepackage{colortbl}

\usepackage[utf8]{inputenc} 
\usepackage[T1]{fontenc}    
\usepackage{hyperref}       
\usepackage{url}            
\usepackage{booktabs}       
\usepackage{nicefrac}       
\usepackage{microtype}      
\usepackage{balance}

\usepackage{xspace}
\usepackage{color}
\usepackage{makecell}

\usepackage{algorithm}
\usepackage{algpseudocode}

\usepackage{subcaption}
\usepackage{url}
\usepackage{cleveref}

\newtheorem{definition}{Definition}

\newcommand{\revise}[1]{\textcolor{black}{#1}}
\newcommand{\MLT}{\textsc{Callisto}\xspace}

\algnewcommand{\LineComment}[1]{\State\(\triangleright\) #1} 
\let\oldReturn\Return
\renewcommand{\Return}{\State\oldReturn}

\def\BibTeX{{\rm B\kern-.05em{\sc i\kern-.025em b}\kern-.08em
    T\kern-.1667em\lower.7ex\hbox{E}\kern-.125emX}}

\author{Sakshi Udeshi,
		Xingbin Jiang,
        Sudipta Chattopadhyay
\IEEEcompsocitemizethanks{
E-mail: sakshi\_udeshi@sutd.edu.sg.}
}

\begin{document}

\title{
Callisto: Entropy based test generation and data quality assessment for Machine Learning Systems\\
}


\maketitle

\begin{abstract} 
Machine Learning (ML) has seen massive progress in the last decade and as a 
result, there is a pressing need for validating ML-based systems. To this end, 
we propose, design and evaluate \MLT -- a novel test generation and data quality 
assessment framework. To the best of our knowledge, \MLT is the first blackbox 
framework to leverage the uncertainty in the prediction and systematically 
generate new test cases for ML classifiers. Our evaluation of \MLT on four 
real world data sets reveals thousands of errors. We also show that leveraging 
the uncertainty in prediction can increase the number of erroneous test cases up 
to a factor of 20, as compared to when no such knowledge is used for testing. 


\MLT has the capability to detect low quality data in the datasets that may 
contain mislabelled data. We conduct and present an extensive user study 
to validate the results of \MLT on identifying low quality data from four 
state-of-the-art real world datasets. 
\end{abstract}


\section{Introduction}
\label{sec:introduction}

Due to the massive progress in Machine Learning (ML) in the last decade, 
its popularity now has reached a variety of application domains, including 
sensitive and safety critical domains, such as automotive, finance, 
education and employment. One of the key reasons to use ML is to automate 
mundane and error-prone manual tasks in decision making. This often results 
in defective machine-learning systems that slip into production run. 
To alleviate this issue, we have developed \MLT, a test generation and data 
quality analysis tool. \MLT leverages the entropy of the outputs of the ML 
classifiers to quantify the uncertainty in the prediction of these 
classifiers. 

To further elucidate our motivation to research this technique, we 
sketch two situations that are likely to occur in the future.  
We also show these problems that may crop up because of the widespread 
proliferation of ML and then we sketch a sample solution for these problems 
that uses our \MLT technique. 

\smallskip
\noindent
\textbf{Test Generation:} It is easy to see that the success of ML is
critically dependent on our ability to collect and annotate data. As 
we move towards more and more sophisticated techniques and tools for 
data collection, we will need to update our toolkit for data management as 
well. A critical part of this ML pipeline remains the testing of 
classifiers produced.

%

\begin{figure}[h]
\begin{center}
\includegraphics[scale=1.1]{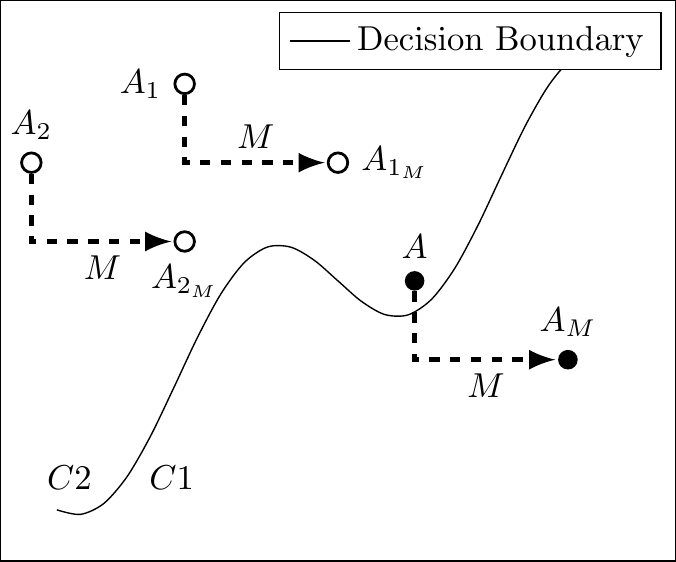}
\end{center}
\caption{Test \MLT intuition}
\label{fig:test-gen-intuition}
\end{figure}

Consider a testing framework which generates tests by applying simple  
transformations such as rotation, zooming and panning to the data in the 
datasets (training and testing datasets). These transformations are 
metamorphic transformations. Let a dataset have $n$ data points.
Assuming the test framework only runs three metamorphic tests, 
a na\"{i}ve approach will have to run $3n$ tests. 
We believe a better approach is to use \MLT to identify inputs 
which are prone to errors. This set is usually much smaller than 
the full dataset and will exhibit highly erroneous behaviours. Test set 
minimisation is a powerful technique to reduce the effort to find 
errors without loss in effectiveness \cite{test-set-min}.  

\MLT employs techniques to quickly and efficiently identify the inputs 
in the datasets for test generation. This aids the user to discover 
erroneous behaviours without extensive testing of the entire dataset. 
We illustrate the intuition behind \MLT's test generation approach 
in \Cref{fig:test-gen-intuition}. Consider a metamorphic transformation 
$M$ (e.g. rotating a picture by a small amount) and inputs $A_1, A_2 $ 
and $A$. A rudimentary approach would be to apply $M$ to all the data 
points leading to large computational overheads. \MLT aims to discover 
points like $A$ which will allow users to selectively apply a metamorphic 
relation to inputs which are likely to cause errors. \MLT will avoid 
points such as $A_1$ and $A_2$.


\smallskip
\noindent
\textbf{Data Quality:} 
%
With the ever-increasing volume of data in training ML systems, it is 
critical to identify the cases where the automatic data generation and/or 
labelling failed and are likely of low quality. 
A na\"{i}ve  approach is to engage humans to verify the data. The cost 
of such an approach is very high and recurring.

Our \MLT approach aims to identify the mislabelled and/or low quality inputs 
for further intervention. The size of these identified sets should be 
significantly smaller in comparison to the full dataset and the cost of 
human intervention is minimal. This approach is first seen in identifying 
wrong labels~\cite{mislabeled-train}. Additionally, we conduct an extensive 
user study to validate our experiments. 
This strategy allows the data to be sanitised efficiently and 
will lead to building of high quality datasets with low computational overhead. 
Consequently, the quality of the classifiers will also improve.

\smallskip
\noindent
\textbf{Solution Sketch:} In a typical ML classifier, there is usually 
a softmax layer~\cite{deeplearning-book}. This layer normalises the 
second last layer's output to a probability distribution of the number 
of output classes. The class with the highest probability is then 
identified as the prediction. The higher the probability, the more 
confident the DNN is in that particular prediction. 

We aim to leverage the output layer of classifiers to make predictions 
about the data point. To illustrate this, consider data points $A$ and 
$B$ with label $l_{2}$ and output vectors, respectively as follows: 
\vspace{-0.05in}
\begin{equation}
\label{eqn:eg1}
\begin{split}
[l_{0}, ~l_{1}, ~l_{2}, ~l_{3}]_A \approx[0.033, 0.033, 0.9, 0.034]
\\
[l_{0}, ~l_{1}, ~l_{2}, ~l_{3}]_B \approx[0.25, 0.2, 0.3, 0.25]
\end{split}
\end{equation}


Intuitively, the quality of the prediction for input $B$, despite being 
correct, is worse. This is because of the lack of high confidence. To 
quantify the confidence of a classifier prediction, we use Shannon Entropy~\cite{informationtheory-book}. 
The Shannon Entropy quantifies 
the diversity of the prediction. It is important to note that the Shannon 
Entropy has been used as a popular measure of diversity in ecological 
literature to quantify the diversity of a 
population~\cite{shannon-ecology}. It is defined as follows: 
\vspace{-0.05in}
\begin{equation}
\label{eqn:shannon}
\begin{split}
H' = - \sum_{i=1}^{N} l_i \ln l_i
\end{split}
\end{equation}
In \Cref{eqn:shannon}, $N$ is the number of labels and $l_i$ is the 
probability that the prediction class belongs to the $i^{th}$ label. 
%
For example, the $H'$ value for input $A$ is $\approx$ 0.44 and the 
same for input $B$ is $\approx$ 1.38. Thus, the higher the Shannon 
index for an output the lower is the quality of the output. 

We use this to build test sets for ML classifiers. The data points 
which have a high Shannon index are the ones likely to be affected most 
by metamorphic transformations.
For Test Generation, we capture the inputs like datapoint $B$ and 
employ  metamorphic relations/perturbations to generate test examples 
efficiently. 

Consider another data point $C$ with label $l_0$. 
Let the prediction output is as follows: 
\vspace{-0.05in}
\begin{equation}
\label{eqn:eg3}
\begin{split}
[l_{0}, ~l_{1}, ~l_{2}, ~l_{3}] \approx[0.0033, 0.0033, 0.99, 0.0034]
\end{split}
\end{equation}

The Shannon Entropy for the output is $\approx$0.066. Thus, the ML model 
is highly confident in the prediction. It is likely that such inputs 
are few in numbers, but it is nevertheless important to find these inputs. 
This is because these data points are the ones which are most likely 
to be mislabelled~\cite{mislabeled-train}. 

The remainder of the paper is organised as follows. After providing a brief 
background and related work (\Cref{sec:background}), we make the following 
contributions:
\begin{enumerate}
	\item We present \MLT, a novel approach to generate the tests for 
	Machine Learning classifiers. We present the first technique, 
	to the best of our knowledge which leverages the entropy 
	of the output of the classifiers for effective test generation. \MLT 
	is also completely blackbox and can be used easily for ML services. 
	(\Cref{sec:methodology})
	
	\item We replicate the results seen before for discovering mislabelled 
	data~\cite{mislabeled-train}. We extend the results to include four 
	real world datasets Fashion-MNIST, 
	CIFAR-10, MNIST and SVHN. (\Cref{sec:results})
	
	\item We conduct an extensive user study to validate the results of our 
	experimentation. Additionally, we publicly release the results of the
	user study. (\Cref{sec:results})
	
	\item We provide the implementation and data of \MLT based on python 
	which is publicly available. (\Cref{sec:results})
\end{enumerate}

We finally discuss lessons learned from building \MLT, threats to its
validity (\Cref{sec:threatsToValidity}) and conclude (\Cref{sec:conclusion})

\section{Background and Related Work}
\label{sec:background}

In recent works, most of the state of the art image classifiers
have been Deep Neural Networks (DNN). As a result of this trend, we 
present some background for DNNs and then move on to testing of ML 
systems. 

\smallskip \noindent
\textbf{Deep Neural Networks:} We can think of a DNN as a function $F$ 
with numerous parameters $F_{\Theta} : \mathbb{R}^N \rightarrow 
\mathbb{R}^M$. The function $F$ with parameters $\Theta$ maps 
an input $x \in \mathbb{R}^N$ to an output $y \in \mathbb{R}^M$. 
Illustrating this with an example, consider an image $x$ (reshaped as a 
vector) that has to be classified into one of $m$ different classes.  
The last layer of the DNN is usually a softmax 
layer~\cite{deeplearning-book} that outputs $y$, which is a vector of
the probabilities of $m$ classes. The predicted label $\hat{y}$ is the 
class with the highest probability: $arg~max_{i \in [1, M]}\ y_{i}$.
In \MLT we use this layer's output to quantify the entropy of a particular 
input as seen in \Cref{eqn:shannon}. 

Internally, the DNN can be understood as a feed forward network. The 
network has $L$ hidden layers which consist of $N_i$ 
neurons where $i \in [1, L]$. These neurons perform computations and the
outputs of these computations are usually called \emph{activations}. For 
the $i^{th}$ layer the vector of activations can be written as follows
\vspace{-0.05in}
\begin{equation} \label{eq:activations}
	a_i = \Delta (w_i \cdot a_{i-1} + b_i)\ ~\forall  i \in [1, L]
\end{equation}

where $a_i \in \mathbb{R}^{N_i}$ and $ \Delta : \mathbb{R}^N \rightarrow 
\mathbb{R}^N$ is an non-linear function.
We can see that $a_0 = x$ and $N_0 = N$. This can be interpreted as 
the inputs to the first layer and the input to the network is the same. 

The parameters $w_i \in \mathbb{R}^{{N}_{i-1}} \times N_i$ and $b_i \in 
\mathbb{R}^{{N}_{i}}$ of \Cref{eq:activations} are fixed weights
and biases respectively and these are learnt during the training phase. 
The output of the network is a function of these activations and can be 
represented as $\gamma(w_{L+1} \cdot a_L + b_{L+1})$, where $\gamma: 
\mathbb{R}^N \rightarrow \mathbb{R}^M$ is usually the softmax layer 
\cite{deeplearning-book}.

%
%

\smallskip\noindent
\textbf{Testing DNN Systems:} One of the 
first works to test DNNs \cite{deeepxplore} presents a whitebox 
differential testing algorithm for systematically finding errors in DNNs.
Another early work~\cite{deeptest} uses a metamorphic testing approach 
to find bugs in DNNs. 
A feature-guided black-box 
approach~\cite{wicker2017feature} was also proposed to validate the safety
of DNNs. 
A set of testing criteria based on multi
level and granularity coverage for testing DNNs is proposed by 
DeepGauge~\cite{deepgauge}. Aequitas~\cite{aequitas} aims to uncover 
fairness violations in machine learning models.
DeepConcolic~\cite{deepconcolic} designs a framework to perform 
concolic testing for discovering robustness violations. A recent work 
\cite{advSampleDetection} uses model mutation methods to detect 
adversarial attacks. PMV \cite{perturbed-model-val} proposes a new 
technique to validate model relevance and detect underfitting or 
overfitting in ML models.

The goal of \MLT is to aid the existing testing systems by minimising the 
test set. 
Specifically, adding tests using all available data is computationally expensive. 
\MLT seeks to make this problem more tractable by constructing a smaller 
set of inputs for testing. Additionally, the inputs should be as effective 
in uncovering low quality data~\cite{mislabeled-train}.

\smallskip\noindent
\textbf{Verification of DNN Systems:}
In contrast to works that attempt verification of DNN systems
\cite{ai2, ReluVal, interval-analysis, reluplex}\MLT aims to preserve the 
flavour of testing in contrast to these 
approaches. In addition to test generation, \MLT also flags data which 
may be of low quality.

\smallskip\noindent
\textbf{Guiding Tests for DNN Systems:}
As testing of ML systems becomes more mainstream, the need to guide these 
massively data intensive systems is apparent. To this end, a recent
work~\cite{surprise-adequacy} proposes a new test coverage metric, 
called \emph{Surprise Adequacy}. This is based on the behaviour of models 
with respect to their training data and develops adequacy criteria for tests. 
Another work~\cite{mislabeled-train} uses entropy to identify low
quality and/or mislabelled data. 

\MLT is the first work, to the best of our knowledge, to use entropy for  
generating tests and minimise test sets. \MLT is also fully blackbox in 
contrast to other test coverage metrics~\cite{surprise-adequacy}. Finally, 
\MLT reimplements an earlier approach~\cite{mislabeled-train}, extends 
the approach to real word datasets and conducts a user study to validate 
the approach, showing that the metric targeted by \MLT can be 
leveraged for multiple use cases in the ML domain.


\section{Methodology}
\label{sec:methodology}

In this section, we elucidate the methodologies behind \MLT in detail. 
\MLT consists of two base algorithms. The first one is the test 
generation framework and another one evaluates data quality. 
Both of these algorithms leverage the Shannon Diversity 
index~\cite{shannon-ecology} to automatically select data points for 
test generation and to evaluate data quality respectively. 

We now introduce some notations that help us to illustrate 
our \MLT approach. These notations are outlined in \Cref{table:notation}.

\begin{table}[H]
	\centering
	\caption{Notations used in \MLT approach}
	\vspace*{-0.1in}
	\label{table:notation}
	\def\arraystretch{1.35}
	\begin{tabular}{|c| p{7cm} |}
	\hline
	$f$ & The machine learning model under test. \\ \hline
	$X$ & The vector of data points. \\ \hline
	$Y$ & The vector of the ground truth labels of the data points. \\ \hline
	$\hat{Y}$ & The vector of the predictions of the data points. \\ \hline
	$\mathbb{S}_{X}$ & The Shannon indices 
	\cite{informationtheory-book} for all $x \in X$ \\ \hline
	
	$\tau_{low}$ & The lower threshold of the Shannon index. Elements 
	with Shannon index lower than $\tau_{low}$ are considered to have 
	very high confidence in their prediction. \\ \hline
	
	$\tau_{high}$ & The higher threshold of the Shannon index. Elements 
	with Shannon index higher than $\tau_{high}$ are considered to have 
	very little confidence in their prediction. \\ \hline
	
	$\delta_{meta}$ & The set of valid metamorphic transformations \\ 
	\hline
	


\end{tabular}
	\vspace*{-0.15in}
\end{table}

\begin{algorithm}[h]
    \caption{\MLT for Test generation}
    {\small
    \begin{algorithmic}[1]  
        \Procedure{Generate}{$X$, $Y$, $\mathbb{S}_{X}$, $\delta_{meta}$, $\tau_{high}$}
        
         	\State $\mathbb{G} \gets \emptyset$ 
            \State $errors \gets \emptyset$ 
            \For{$s_{x_i} \in \mathbb{S}_{X}$}
            	\If{$s_{x_i} > \tau_{high}$ and $y_i == \hat{y}_i$}
            		\State $\mathbb{G} \gets \mathbb{G} \cup \{x_i\}$
            	\EndIf
            \EndFor
            
            \For{$x_i \in \mathbb{G}$}
            		\LineComment Choose a transformation to 
            		apply
            		\State {$T \gets $ {\tt Choice}
            		($\delta_{meta}$)}
            		\State $x^T_i \gets $ {\tt Apply\_Transform} ($x_i, 
            		T$)
            	
            	\If{$f(x^T_i) \neq y_i$}
            		\State $errors \gets errors \cup \{x^T_i\}$
            	\EndIf
            \EndFor
                        
            \Return $errors$
        \EndProcedure
    \end{algorithmic}}
    \label{alg:test-generation}
\end{algorithm}

\begin{figure}[h]
\begin{center}
\includegraphics[scale=0.5]{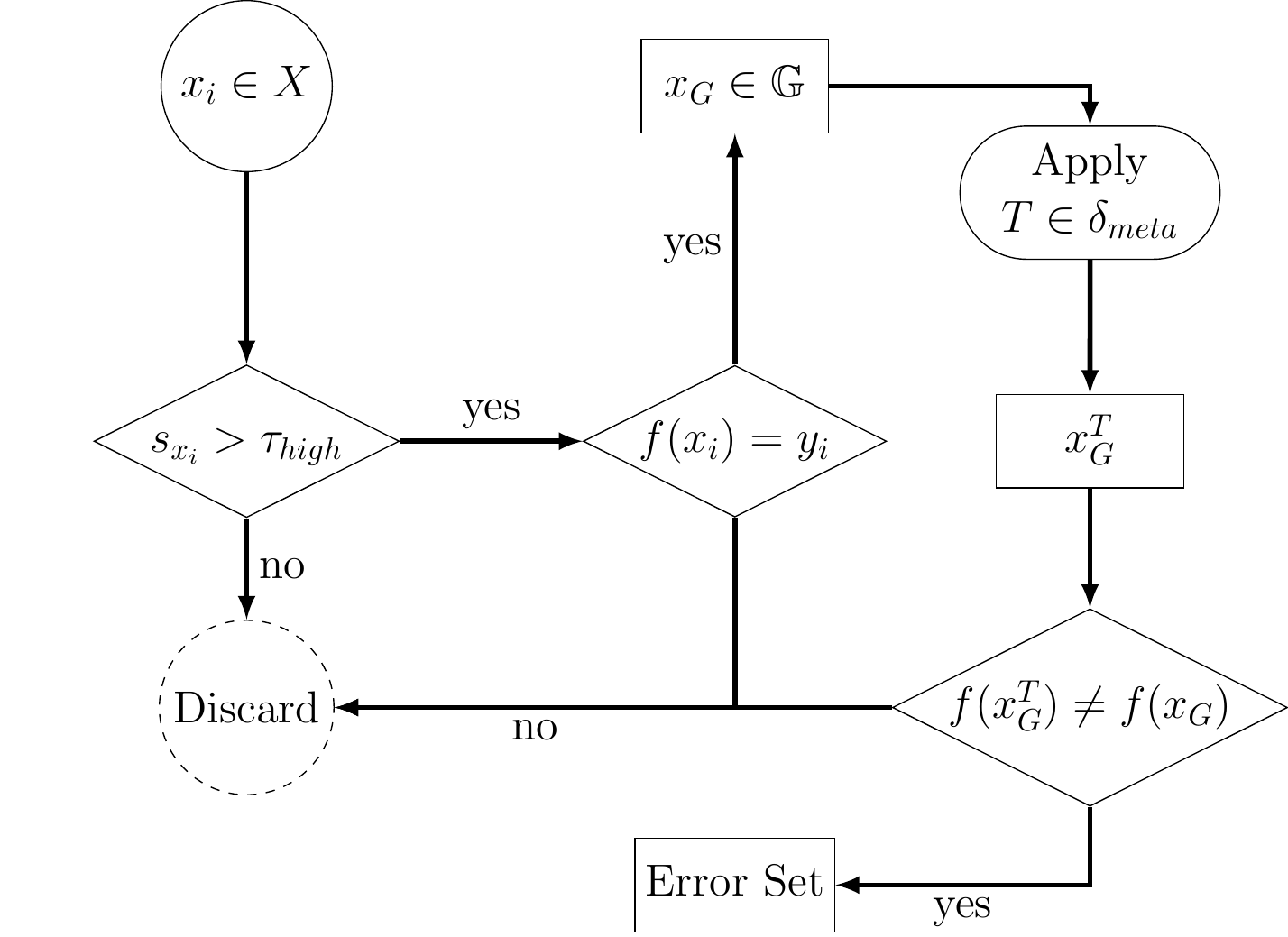}
\end{center}
\caption{Test generation with \MLT}
\label{fig:test-generation-block-diagram}
\end{figure}

\smallskip\noindent
\textbf{Test generation in \MLT:} \Cref{alg:test-generation} and 
\Cref{fig:test-generation-block-diagram} outline the overall test generation 
process that \MLT implements. The input is a large corpus of data points 
(usually the training data and the testing data) $X$ and the 
labels of the training data $Y$. The training data set is generally 
very large and it is computationally expensive to generate and 
evaluate metamorphic tests for all $x_i \in X$. 

To aid efficient test generation, \MLT generates tests only for those data 
points where it is likely for the metamorphic transformations to cause an 
error. Such data points are the ones whose output $f(x_i)$ has a Shannon 
index $s_i \in \mathbb{S}_{X}$ such that $s_i > \tau_{high}$. Intuitively, 
these data points indicate scenarios where the model $f$ had low confidence 
in the prediction. 
We also impose the additional condition that the output $f(x_i)$ is 
equal to the label $y_i \in Y$. We construct a set $\mathbb{G}$ that 
contains data points satisfying these conditions. 
It is important to note that these are clean inputs and not adversarially 
crafted inputs where the confidence is artificially high. As a result, we 
expect a well trained model to be confident of the output and low confidence 
is an indication of a gap in learning~\cite{mislabeled-train}.

Once the set $\mathbb{G}$ is constructed, \MLT iterates over each 
datapoint in this set and recursively applies a pre-selected 
metamorphic transformation. 
For example, consider a data point $x_{G} \in \mathbb{G}$ where we 
pick a transformation 
$T \in \delta_{meta}$ and apply it to $x_{G}$ to produce $x_{G}^T$.
Once we have this input $x_{G}^T$, we 
add $x_{G}^T$ to the error set if and only if $f(x_{G}^T) \neq f(x_{G})$. 
\MLT then returns the error set for the user to evaluate.
We summarise this approach in \Cref{fig:test-generation-block-diagram}. 


\begin{algorithm}[h]
    \caption{\MLT for Low Data Quality Detection}
    {\small
    \begin{algorithmic}[1]  
        \Procedure{Detect}{$X$, $Y$, $\mathbb{S}_{X}$, $\tau_{low}$}
        	\State $\mathbb{F} \gets \phi$
            \For{$s_{x_i} \in \mathbb{S}_{X}$}
            	\If{$s_{x_i} < \tau_{low}$ and $y_i \neq \hat{y}_i$}
            		\State $\mathbb{F} \gets \mathbb{F} \cup \{x_i\}$
            	\EndIf
            \EndFor
        
        	\Return $\mathbb{F}$
        \EndProcedure
    \end{algorithmic}}
    \label{alg:data-quality}
\end{algorithm}

\smallskip\noindent
\textbf{Identifying low quality data in \MLT:} The aim of this part of 
\MLT is to create a set $\mathbb{F}$ which flags and identifies potential 
low quality data. In an annotated dataset which is used for supervised 
learning, it is possible that some data is mislabelled and/or ambiguous. 
Manually going through this data incurs high cost. We formalise the 
notion of low quality data below. 

\theoremstyle{definition}
\begin{definition}{\textbf{(Low Quality Data)}}
\label{def:low-quality-data}
{
Consider a data point $x \in X$ with label $y \in Y$ and prediction 
$\hat{y} \in \hat{Y}$. We consider the data point $x$ to be of low 
quality if $y$ is not the correct label and instead $\hat{y}$ is the 
}
\end{definition}


\Cref{alg:data-quality} illustrates the \MLT approach to discover low quality 
data. For each input $x_i \in X$, we check the Shannon index $s_{x_i} \in \mathbb{S}_X$. 
A Shannon index lower than the threshold $\tau_{low}$ can be interpreted
as the model $f$ having high confidence in the prediction. \MLT 
also requires that the predicted value $\hat{y}_i$ to be different than 
the label $y_i$. Intuitively, this means that the prediction for data 
point $x_i$ has high confidence, yet the prediction does not match with 
the label. Thus, it is likely that the label $y_i$  is incorrect and 
therefore, $x_i$ is considered to be of low quality. As 
in the previous section, it is important to note that $x_i$  is a 
clean input and not an adversarially crafted input where the confidence 
is artificially high.

\begin{figure}[h]
\begin{center}
\begin{tabular}{ccc}
\includegraphics[scale=0.3]{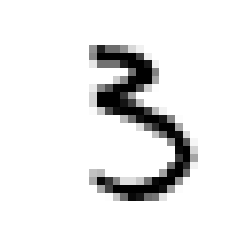} & 
\includegraphics[scale=0.3]{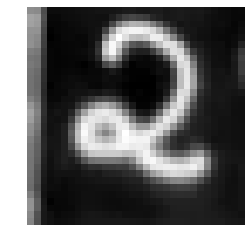} & 
\includegraphics[scale=0.3]{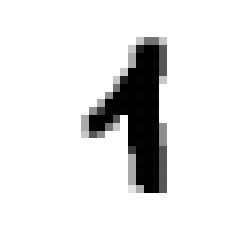}\\
{\makecell{ Label - 5 \\ Prediction - 3 \\ (a)}} & {\makecell{Label - 6 \\ Prediction - 2\\ (b)}} & {\makecell{ Label - 4 \\ Prediction - 1\\ (c)}}\\
\end{tabular}
\end{center}
\caption{Some examples of low quality data found in the MNIST~\cite{mnist} and SVHN~\cite{mnist} datasets}

\label{fig:low-data-quality-examples}
\end{figure}

Some of the examples of low quality data found by \MLT can be seen in 
\Cref{fig:low-data-quality-examples}.



To further validate the efficacy of \MLT, we have designed a user study. 
The objective of this user study is to check whether the low quality data 
points discovered by \MLT are indeed of low quality perceived by users. 
To design the user study, we constructed a  set of representative examples 
of the low quality data found in the MNIST Digit~\cite{mnist}, Fashion-MNIST~\cite{fashion-mnist}, 
CIFAR-10~\cite{cifar-10} and SVHN~\cite{SVHN} datasets. 
These examples were presented to users and they were asked to choose 
the correct output class of an example $x$ between two options. The first 
option was the respective prediction $\hat{y}$ or the second option was 
the label $y$. Of course, the sequence in which these options appear 
was randomized for each user. We also asked each user to rate the confidence 
of her choice on a five point Likert Scale~\cite{likert-scale}. 
Examples of the questions that we asked for \Cref{fig:low-data-quality-examples}(a) 
are as follows:  

\begin{verbatim}
   What is the number displayed above? 
          Option a - 5

          Option b - 3
\end{verbatim}


\begin{verbatim}
How confident are you in your answer? 
          1  2  3  4  5
 Low Confidence     High Confidence
\end{verbatim} 
We elaborate on the results of this user study in \Cref{sec:results}.

\section{Results}
\label{sec:results}

To evaluate the efficacy of \MLT, we answer the following research questions.

\begin{table}
\centering
\arrayrulecolor{black}
\caption{Test Generation Effectiveness}
\label{table:rq1-1}
\begin{tabular}{!{\color{black}\vrule}c!{\color{black}\vrule}c!{\color{black}\vrule}c!{\color{black}\vrule}c!{\color{black}\vrule}c!{\color{black}\vrule}} 
\cline{2-5}
\multicolumn{1}{l!{\color{black}\vrule}}{} & \multicolumn{4}{c!{\color{black}\vrule}}{Ratio of erroneous inputs (\#error/\#test)}  \\ 
\hline
Shannon Threshold                          & Panning & 2D rotation & Affine & Perspective           \\ 
\hline
\multicolumn{5}{!{\color{black}\vrule}c!{\color{black}\vrule}}{Fashion MNIST}                       \\ 
\hline
$s_x < 0.001 $ & 0.19    & 0.58        & 0.41   & 0.07                  \\ 
\hline
$s_x > 0.4 $ & 0.61    & 0.78        & 0.80   & 0.58                  \\ 
\hline
\multicolumn{5}{!{\color{black}\vrule}c!{\color{black}\vrule}}{MNIST-Digit}                         \\ 
\hline
$s_x < 0.001$& 0.20    & 0.12        & 0.07   & 0.03                  \\ 
\hline
$s_x > 0.4 $ & 0.65    & 0.51        & 0.45   & 0.51                  \\ 
\hline
\multicolumn{5}{!{\color{black}\vrule}c!{\color{black}\vrule}}{CIFAR-10}                            \\ 
\hline
$s_x < 0.001$& 0.09    & 0.55        & 0.53   & 0.27                  \\ 
\hline
$s_x > 0.4$& 0.40    & 0.62        & 0.63   & 0.51                  \\ 
\hline
\multicolumn{5}{!{\color{black}\vrule}c!{\color{black}\vrule}}{SVHN}                                \\ 
\hline
$s_x < 0.001$& 0.03    & 0.46        & 0.59   & 0.14                  \\ 
\hline
$s_x > 0.4$ & 0.54    & 0.79        & 0.84   & 0.58                  \\
\hline
\end{tabular}
\arrayrulecolor{black}
\end{table}

\begin{figure*}[h]
\begin{center}
\begin{tabular}{ccccc}
\includegraphics[scale=0.3]{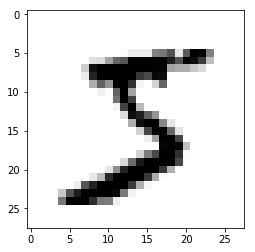} & 
\includegraphics[scale=0.3]{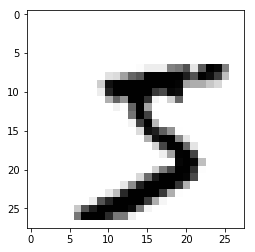} & 
\includegraphics[scale=0.3]{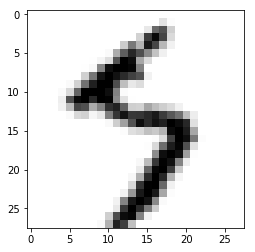} & 
\includegraphics[scale=0.3]{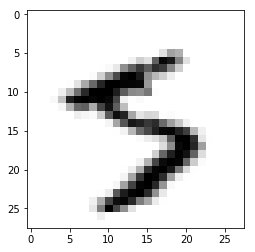} &
\includegraphics[scale=0.3]{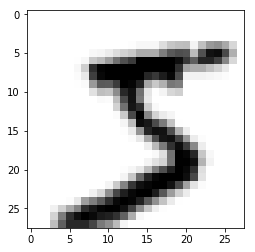}\\
{\makecell{Original \\(a)}} &{\makecell{Panning \\(b)}} & {\makecell{2D Rotation \\(b)}} & {\makecell{Affine \\(d)}}  & {\makecell{Perspective \\(e)}}\\
\end{tabular}
\end{center}
\caption{Transformations of an image}

\label{fig:image-transformation}
\end{figure*}

\begin{center}
\begin{tcolorbox}[width=\columnwidth, colback=gray!25,arc=0pt,auto outer arc]
\textbf{RQ1.1: Is the test generation effective?}
\end{tcolorbox}
\end{center}
\vspace*{-0.05in}

To evaluate the efficiency of this research question, we compute the ratio of 
erroneous inputs obtained via metamorphic transformations. We say a transformed 
input is an error when the prediction of the model does not match the corresponding 
label. In our evaluation, we picked four transformations, 
namely panning, 2D rotation, affine and perspective. These transformations 
can be seen in \Cref{fig:image-transformation}. It has been shown that 
Machine Learning models are not robust to even simple transformations 
\cite{robust-adversarial-examples}. The objective for Callisto is to 
minimise the number of transformations and maximise the number of erroneous 
inputs. We show that the Shannon index is an effective measure to maximise 
the error rates. Inputs with a high Shannon index (more than 0.4), as seen in 
\Cref{table:rq1-1}, consistently show higher error rates in comparison to 
the entire dataset (Shannon threshold - 0.0 ). Moreover, the effectiveness 
of the test generation does not depend on the type of transformation. All 
four of the transformations show an increase in the ratio of erroneous inputs with 
a higher Shannon index.  

\begin{center}
\begin{tcolorbox}[width=\columnwidth, colback=gray!25,arc=0pt,auto outer arc]
\textbf{RQ1.2: How do the error rates vary with Shannon index?}
\end{tcolorbox}
\end{center}
\vspace*{-0.05in}

\begin{figure}[h]
\begin{center}
\includegraphics[scale=0.25]{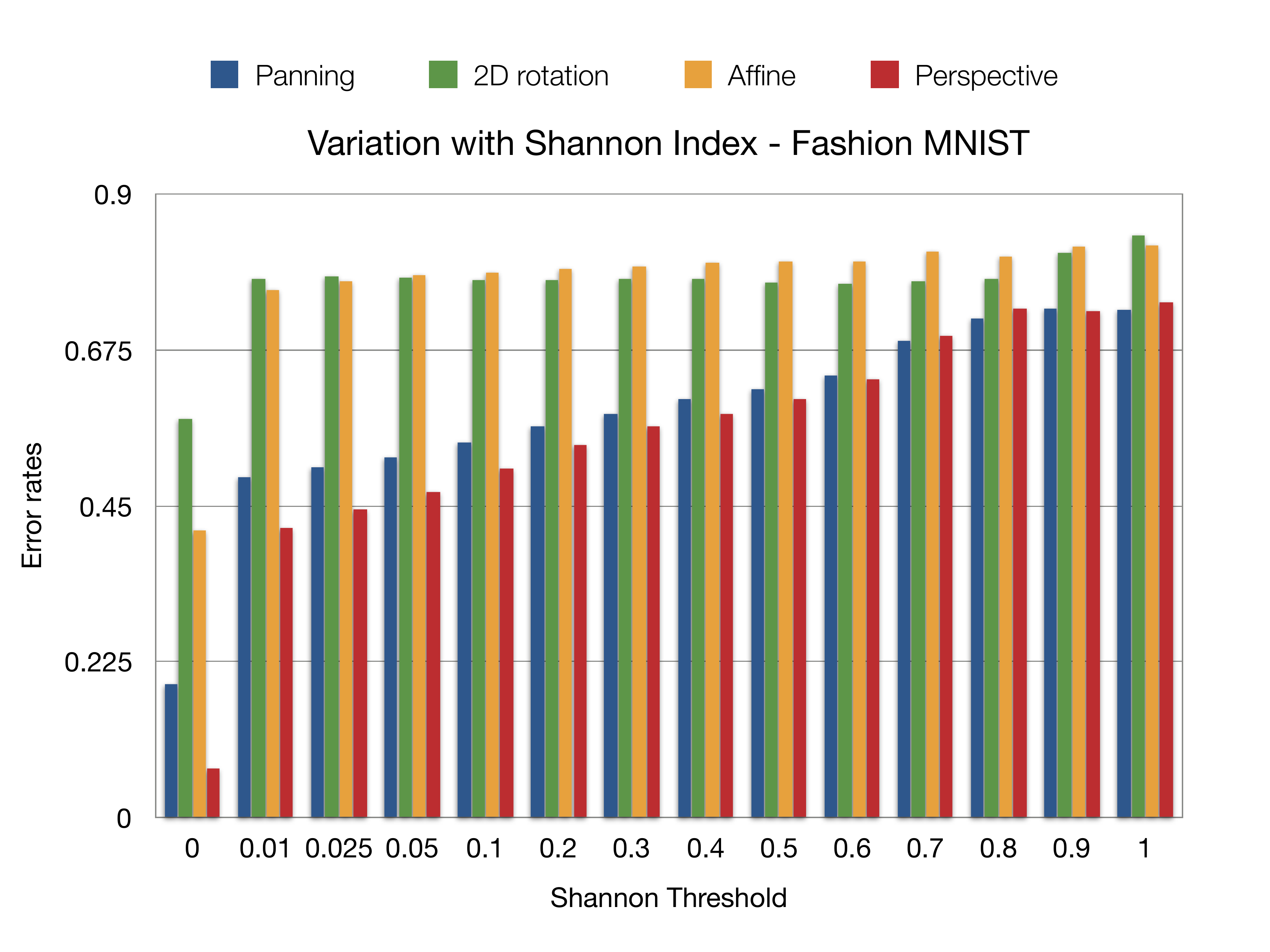}
\end{center}
\caption{Test generation with \MLT}
\label{fig:threshold-vary}
\end{figure}

To answer this research question, we evaluated the error rates for the 
Fashion MNIST as seen in \Cref{fig:threshold-vary}. Intuitively, the 
error rates should increase with an increase in the Shannon Threshold. 
This is because the model outputs that have a high Shannon index can 
be understood as being less confident in their prediction and being 
more brittle. Thus the respective inputs are more prone to errors due 
to transformations. 

As we can see in \Cref{fig:threshold-vary}, our intuition holds. An 
increase in the Shannon threshold causes an increase in the error 
rates across all the transformations. For example, the error rate 
increased up to almost ten times in the case of the perspective 
transformation. \revise{This trend holds for all the 
datasets, please refer to the supplementary materials (\Cref{sec:conclusion}).} 

\begin{center}
\begin{tcolorbox}[width=\columnwidth, colback=gray!25,arc=0pt,auto outer arc]
\textbf{RQ2: Is low quality data effectively identified?}
\end{tcolorbox}
\end{center}
\vspace*{-0.05in}

\begin{table}
\centering

\caption{Identifying low quality data}
\label{table:rq2-1}
\arrayrulecolor{black}
\begin{tabular}{!{\color{black}\vrule}c!{\color{black}\vrule}c!{\color{black}\vrule}c!{\color{black}\vrule}c!{\color{black}\vrule}} 
\hline
\multicolumn{1}{!{\color{black}\vrule}l!{\color{black}\vrule}}{} & Image \#1                                            & Image \#2                                            & Image \#3                                             \\ 
\hline
\multicolumn{4}{!{\color{black}\vrule}c!{\color{black}\vrule}}{Fashion MNIST}                                                                                                                                        \\ 
\hline
\%users choosing prediction                                      & \multicolumn{1}{c!{\color{black}\vrule}}{95.5} & \multicolumn{1}{c!{\color{black}\vrule}}{94.5} & 83.90                                           \\ 
\hline
Confidence  (1 to 5)                                                     & 4.17                                           & 3.71                                           & 4.40                                            \\ 
\hline
\multicolumn{4}{!{\color{black}\vrule}c!{\color{black}\vrule}}{MNIST-Digit}                                                                                                                                          \\ 
\hline
\%users choosing prediction                                      & 98                                             & 100                                            & 90.50                                           \\ 
\hline
Confidence   (1 to 5)                                                    & 4.54                                           & 4.76                                           & 4.39                                            \\ 
\hline
\multicolumn{4}{!{\color{black}\vrule}c!{\color{black}\vrule}}{CIFAR-10}                                                                                                                                             \\ 
\hline
\%users choosing prediction                                      & \multicolumn{1}{c!{\color{black}\vrule}}{32.7} & \multicolumn{1}{c!{\color{black}\vrule}}{27.1} & \multicolumn{1}{l!{\color{black}\vrule}}{85.9}  \\ 
\hline
Confidence (1 to 5)                                                  & 3.77                                           & 3.34                                           & 3.32                                            \\ 
\hline
\multicolumn{4}{!{\color{black}\vrule}c!{\color{black}\vrule}}{SVHN}                                                                                                                                                 \\ 
\hline
\%users choosing prediction                                      & 100                                            & 100                                            & 100                                             \\ 
\hline
Confidence   (1 to 5)                                                   & 4.74                                           & 4.82                                           & 4.92                                            \\
\hline
\end{tabular}
\arrayrulecolor{black}
\end{table}

The usage of the Shannon diversity to identify mislabelled data was first 
proposed to detect mislabelled training instances~\cite{mislabeled-train}. 
Our solution to detect low quality data (i.e. \Cref{alg:data-quality}) is 
similar, but we conduct a thorough user study to validate our results.  

We conducted a survey with representative examples from the MNIST-Digit, 
Fashion-MNIST, SVHN and CIFAR-10 datasets. 
These examples had incorrect predictions and the prediction outputs had low 
Shannon index. Our intuition is that these are the images which are likely 
of low quality. To evaluate this, we conducted a survey on 
the Amazon's mTurk~\cite{amazon-mTurk} \revise{with 197 users}. We asked 
the users to choose between the label and the predicted output for three 
representative examples (from each dataset) which were predicted 
incorrectly, but had a low Shannon index (i.e. high output confidence). 

As seen in \Cref{table:rq2-1}, for the Fashion MNIST, MNIST-Digit and SVHN 
the prediction were largely chosen over the label and the confidence is 
generally high. For CIFAR-10, in two out of the three examples, the label 
was chosen as a majority. \revise{We believe that the low confidence across 
CIFAR indicates that these images are generally hard to label and 
understand. 
Nevertheless, we acknowledge that further investigations are required to 
detect low quality data in datasets like CIFAR.} 



\section{Threats to Validity}
\label{sec:threatsToValidity}

\noindent \smallskip
\textbf{Testing Vision Systems:} The test subjects for \MLT are
exclusively vision systems. We have not validated \MLT on other 
classifiers, such as text classifiers. For other types of ML systems, 
\MLT might need to involve additional metamorphic transformations. 
Nonetheless, we believe that Shannon Entropy, as leveraged by \MLT, 
is a general property that can easily be used by other 
types of classifiers. 

\noindent \smallskip
\textbf{CIFAR-10 Data Quality:} For \MLT we have reimplemented the approach 
first seen for mislabelling datasets~\cite{mislabeled-train}. This approach 
does not perform well for the CIFAR-10 dataset in two out of the three 
representative images as seen in \Cref{table:rq2-1}. Further investigation 
is necessary along this line of research. 
We aim to investigate this for future work.

\section{Conclusion}
\label{sec:conclusion}

In this paper, we present \MLT, a novel entropy-based test generation 
framework. We show that using existing information about the output 
uncertainty, we can effectively generate erroneous inputs. 
We also reimplement and validate an approach first seen for finding 
mislabelled data~\cite{mislabeled-train}. Additionally, we have 
conducted extensive user studies to try and validate the results 
seen in \MLT. 

\MLT is a major step towards pushing the state-of-the-art in testing 
of ML models, which bring along several fresh challenges due 
to their unique nature. \MLT is completely blackbox and does not require 
any information about the structure of classifiers. Therefore, \MLT can 
easily be deployed for testing ML services. To promote research in this area 
and reproduce our results, we have made our implementation and all 
experimental data publicly available:

\begin{center}
\url{https://github.com/sakshiudeshi/Callisto}
\end{center}

\bibliographystyle{plainurl}
\bibliography{Callisto}

\end{document}